\documentclass[a4paper,conference]{IEEEtran}
\usepackage{times}
\usepackage{soul}
\usepackage{url}
\usepackage[hidelinks]{hyperref}
\usepackage[utf8]{inputenc}
\usepackage[small]{caption}
\usepackage{graphicx}
\usepackage{cite}
\usepackage{amsthm}
\usepackage{booktabs}
\usepackage{algorithm}
\usepackage{algorithmic}
\usepackage{amsmath,amssymb,amsfonts}
\usepackage{graphicx}
\usepackage{textcomp}
\usepackage{xcolor}
\usepackage{hyperref}       
\usepackage{url}            
\usepackage{amsfonts}       
\usepackage{nicefrac}       
\usepackage{microtype}      
\usepackage{mathrsfs}
\usepackage{amsmath}
\usepackage{amssymb}
\usepackage{graphicx, subfig}
\usepackage{epstopdf}
\usepackage{tikz}
\usepackage{pgf}
\usepackage{multirow}
\usepackage{pgfplots}
\usepackage{float}
\usepackage{pifont}
\newcommand{\cmark}{\ding{51}}%
\newcommand{\xmark}{\ding{55}}%

\begin{document}
%
\title{Learning to Prune in Training via Dynamic Channel Propagation}



%
\author{\IEEEauthorblockN{Shibo Shen\IEEEauthorrefmark{1},
Rongpeng Li\IEEEauthorrefmark{1},
Zhifeng Zhao\IEEEauthorrefmark{2},
Honggang Zhang\IEEEauthorrefmark{1}, and
Yugeng Zhou\IEEEauthorrefmark{3}}
\IEEEauthorblockA{\IEEEauthorrefmark{1}College of Information Science and Electronic Engineering, Zhejiang University, Hangzhou 310027, China\\ Email: \{shenshibo, lirongpeng, honggangzhang\}@zju.edu.cn}
\IEEEauthorblockA{\IEEEauthorrefmark{2}Zhejiang Lab, Hangzhou 311121, China}
\IEEEauthorblockA{\IEEEauthorrefmark{3}Zhejiang Wanfeng Technology Development Company, Ltd., Zhejiang 312500, China}
}

\maketitle

\begin{abstract}
In this paper, we propose a novel network training mechanism called "dynamic channel propagation" to prune the neural networks during the training period. In particular, we pick up a specific group of channels in each convolutional layer to participate in the forward propagation in training time according to the significance level of channel, which is defined as channel utility. The utility values with respect to all selected channels are updated simultaneously with the error back-propagation process and will adaptively change. Furthermore, when the training ends, channels with high utility values are retained whereas those with low utility values are discarded. Hence, our proposed scheme trains and prunes neural networks simultaneously. We empirically evaluate our novel training scheme on various representative benchmark datasets and advanced convolutional neural network (CNN) architectures, including VGGNet and ResNet. The experiment results verify the superior performance and robust effectiveness of our approach\footnote{Our code is available at \url{https://github.com/shibo-shen/Dynamic-Channel-Propagation}}.
\end{abstract}


%
\IEEEpeerreviewmaketitle

\section{Introduction}
It is generally believed that the ability of Convolutional Neural Network (CNN) depends on its depth \cite{he2016residual,simonyan15very,szegedy2015going}. However, deep CNN leads to  huge overhead in both storage and computing. In order to make deep learning available on low-power devices, such as mobile phones, it is necessary to reduce its huge complexity. Among many compression methods, a kind of strategy called channel pruning is specifically designed to reduce complexity of the convolutional layer \cite{li2017pruning,he2017channel,molchanov17,liu2017learning}.

\begin{figure}
	\centering
	\includegraphics[scale=0.35]{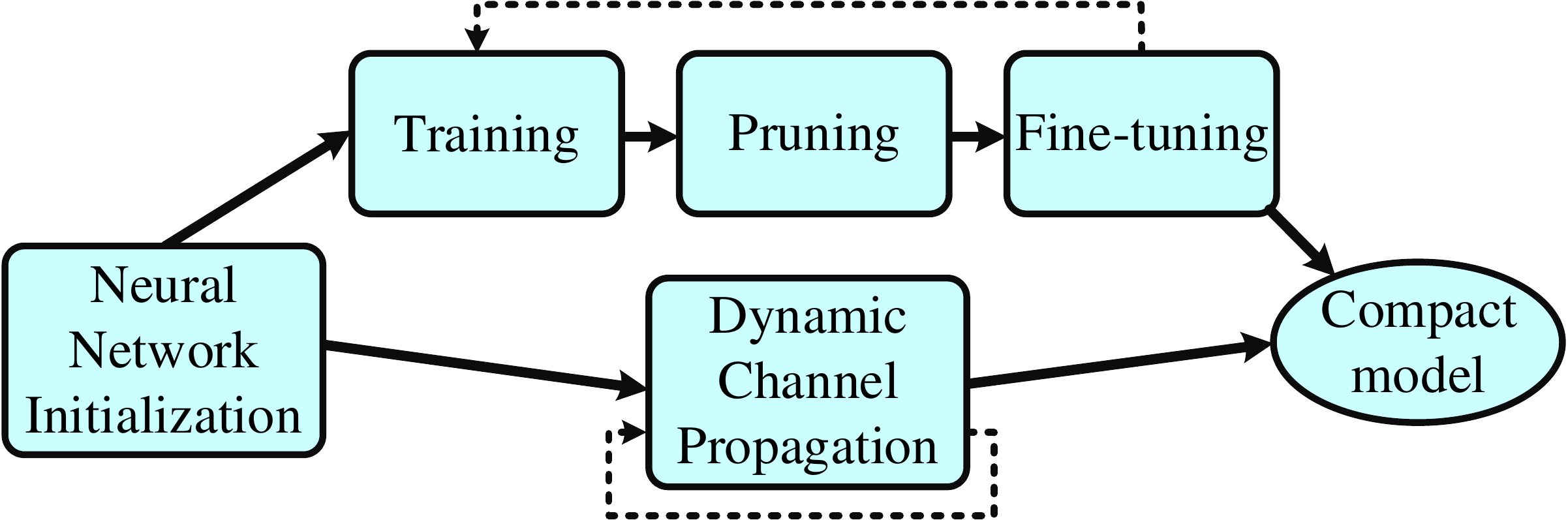} 
	\caption{The comparison of traditional pruning procedure and our proposed approach. The dash lines represents iterative pruning to achieve a high compression rate.}
	\label{f1}
\end{figure}

Channel pruning has been widely studied and achieves exciting results \cite{he2018soft,luo2017ThiNet,li2017pruning,he2017channel,molchanov17,liu2017learning}. One drawback is that neither of these methods escape a three-stage procedure, i.e., training a redundant network, pruning it and fine-tuning for accuracy recovery as depicted in \figurename{\ref{f1}}, which is very cumbersome and whose performance is heavily dependent on the pre-trained model. In fact, recent research shows that pruning based on a well-trained model performs no better than training a pruned network from scratch \cite{liu2018rethinking}. Thus in order to simplify the traditional pruning procedure, we propose to prune the neural network in training by dynamically propagating the channels with large significance evaluation, which is based on channel utility characteristics. 

We learn from both Dropout \cite{srivastava2014dropout} and channel-wise Dropout \cite{tompson2015efficient} that only part of channels are selected to retain their values at a forward iteration while other channels are zeroed out. Due to that approach the chosen channels at each feed-forward iteration are different, one may regard the training process with dropout as a training integration of multiple sub-neural-networks. However, the whole network channels are activated at inference time. We accordingly modify the dropout method that at training time, the activated channels selection is no longer random but dependent on their channel utilities, which are defined as channels' sensitivity to the network loss. During the testing period, channels which are activated in the training stage still remain their values whereas those zeroed-out channels are pruned. 

\begin{figure}[t]
	\centering
	\includegraphics[scale=0.58]{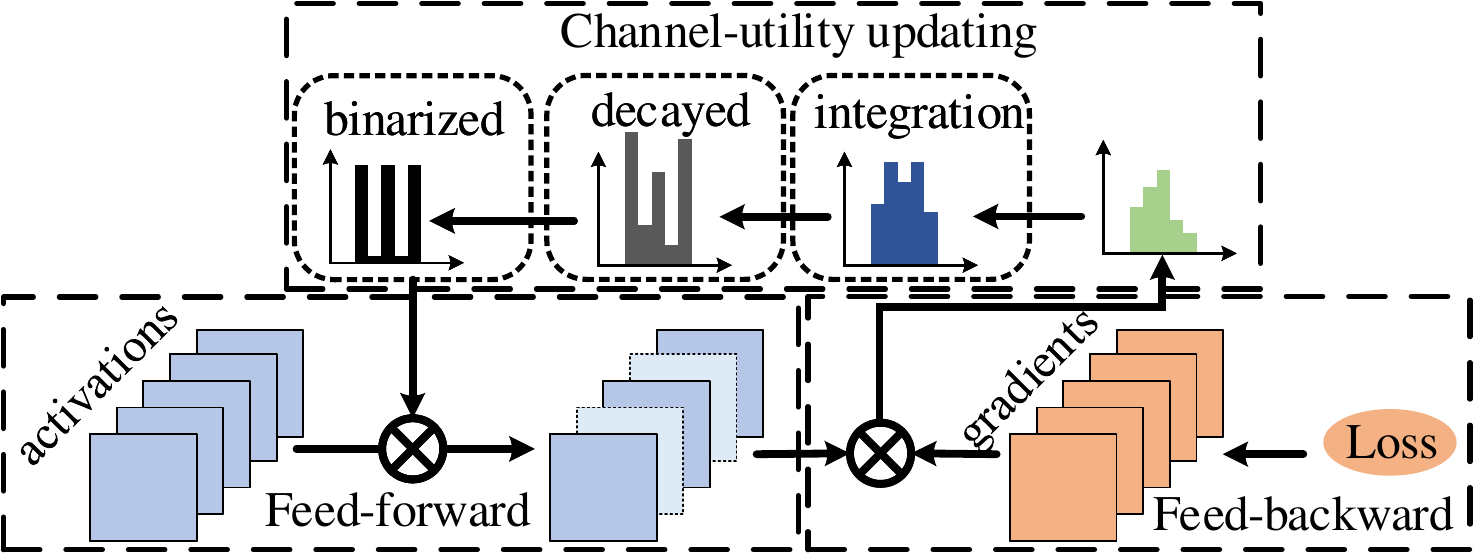} 
	\caption{A high level view of the proposed training process in forward and backward pass.}
	\label{f2}
\end{figure}

Our proposed scheme contains channel selection and utility updating in the forward and backward pass, respectively. As shown in \figurename{\ref{f2}}, those channels with larger utility values are selected to be "activated", in which they maintain activation values in forward propagation while others' activation are set to zero. In the backward pass with the error propagation, we can calculate the update amount of the chosen activated channel utilities using the criterion suggested by \cite{molchanov17}. This dynamic adjustment process continues until the training ends. Finally, we prune channels with small utility values meanwhile set up corresponding filters to implement model compression and acceleration. Hence, our proposed approach can train and prune neural networks at the same time. In other words, our approach can directly transform a redundant, ambiguous neural network into a compact, expressive one.

Intuitively, our goal is to pick up a group of channels which constitutes the sub-neural-architecture that performs better than other sub-neural-networks. In addition, traditional methods prune channels on the basis of a well-trained complex model. However, there is no pre-trained model in some scenarios and the previous three-step model training procedure, including pre-training, pruning and fine-tuning, is not cost-effective. Note that the paper \cite{ding2019global} presents similar ideas with (1) end-to-end training; (2) no need for a time-consuming re-training process after pruning. We believe our proposed scheme implements the same functions in a totally different way.

Our major contributions are summarized as follows. We incorporate the traditional pruning process into the network training and propose a novel training mechanism. Our algorithm is light-weight and can be easily incorporated into training processes of any neural network. As long as the pruning rate is given, our scheme can unveil a sub-neural-architecture itself and learn the weights simultaneously. Extensive experiments are carried out to show the superior performance of our scheme, including pruning VGG \cite{simonyan15very} on CIFAR-10 \cite{alex2009learning} and pruning ResNet \cite{he2016residual} on CIFAR-10 and ILSVRC-12 \cite{jia2009imagenet}.
\section{Related Work}
\textbf{Channel pruning}. Channel pruning has been widely studied to compress convolutional layers in deep neural networks. Existing pruning approaches can be divided into dynamic strategy \cite{gao2019dynamic, lin2017runtime} and static strategy \cite{wen2016Learning,li2017pruning,luo2017ThiNet,liu2017learning,molchanov17,he2017channel}. The dynamic strategy preserves all channels but dynamicly chooses a part of channels during an inference time while the static strategy directly removes less important channels. Among dynamic approaches, the work \cite{lin2017runtime} designs a recurrent neural network (RNN) to encode the previous layers' information, which is applied to predict the current useful filters, thus avoiding unnecessary convolutions. Such decision of recurrent network is adjusted by reinforcement learning rather than directly using the gradients. The paper \cite{gao2019dynamic} introduces a simpler channels decision algorithm that considers the adjacent layers correlation by constructing a predictor and further makes the predictor trainable with parameters in the neural network.

Different from dynamic strategies, the static strategy generates a more compact model for inference directly and averts the extra overhead in channel prediction. The work \cite{wen2016Learning} adds an extra loss function called group LASSO to learn the sparsity in filters. This process reduces the magnitude of filters and forces some of them to zero values, which can be removed at the end of training. In addition, the paper \cite{liu2017learning} uses group LASSO to learn the sparsity of weights in the batch normalization \cite{serpy2015batch} layers following the convolutional layers. Such weights are multiplied with the corresponding channels and therefore can be regarded as a measure of channel sparsity. Different from learning the sparsity during training period, the authors of \cite{he2017channel} propose to learn the sparsity on the basis of a well-trained model by optimizing the formulation of reconstruction error. Moreover the paper \cite{luo2017ThiNet} also focuses on the reconstruction error in which they sample part of all input activation (called sliding window) to minimize the gap between the output of overall and sampled activation. 

Some other approaches prune channels through measuring the criterion of channel significance \cite{li2017pruning,molchanov17}. \cite{li2017pruning} prunes channels based on their corresponding filters' importance evaluation as $\mathcal{L}_1$ or $\mathcal{L}_2$ norm.
They believe that small-norm filters generate less important channels which can be safely removed. The work \cite{molchanov17} introduces another criterion to evaluate channel saliency based on Taylor expansion over the cost function while the specific channel is removed. The channel significance in terms of these criteria is gradually accumulated during training phase and finally ranked. In addition, \cite{zhuang2019discrimination} carefully designs a discrimination-aware loss to keep channels that really contribute to the discriminative power of the network and prune the others.

Different from the aforementioned pruning approaches, which all depend on a three-stage procedure, both \cite{he2018progressive} and \cite{he2018soft} propose to update the pruned filters while training the model, called Soft Filter Pruning (SFP), whose idea is similar to ours. However, SFP prunes filters when a training epoch is over and needs extra reconstruction process to restore the pruned filters, which basically is different from our scheme that selects channels at each mini-batch forward iteration (without pruning other channels) and updates the utility of all selected channels in the backward pass. Intuitively, our scheme is smoother in that the pruned filters’ values are preserved, different from SFP where the filters are roughly zeroed out, making it more stable for network convergence. In addition to pruning schemes based on existing DNN models, recent works \cite{Prabhu_2018_ECCV, mcdonaldSSRN2019} propose to 
construct sparsely connected DNNs based on considering the adjacent layers as a sparse bipartite graph whose edges between input and output neurons are much less than its densely connected counterpart, and to train such neural networks from scratch. Their algorithm can be seen as a supplement to the pruning-based scheme for DNN compression.

\textbf{Channel-wise dropout}. One may consider our proposed scheme as a combination of channel pruning and dropout \cite{hou2019weighted,tompson2015efficient} where only part of channels are allowed to participate in the forward propagation in the training phase, but in the testing phase, all channels are activated. Channel-wise dropout can effectively reduce the similarity of different channels, which is argued by \cite{tompson2015efficient}. Weighted Channel Dropout \cite{hou2019weighted} selects forward channels according to their activation values, which is similar to some criteria used in channel pruning \cite{li2017pruning}. Inspired by such approach, we design an end-to-end strategy that only a set of effective channels are selected to become activated at every training iteration and when the training ends, we can obtain a more compact network by directly copying the corresponding effective filters without extra fine-tuning. 
\section{Dynamic Channel Propagation}
\subsection{Notations}
Before illustrating the proposed mechanism in detail, we formally introduce some symbols in this sub-section. Suppose we have a deep neural network with $L$ convolutional layers. We use $\mathbf{w}^l_{k}$ and $\mathbf{z}^l_{k}$ to represent the convolutional kernel and the individual output channel of $l$th convolutional layer, respectively. The index $k$ in the subscript means $k$th channel of the corresponding output.  We can see that $\mathbf{w}^l_k \in \mathbb{R}^{C^{l-1}\times K\times K}$ and $\mathbf{z}^l_k \in \mathbb{R}^{H^l\times W^l}$ where $C^{l-1}$ represents the number of input channels to the $l$th layer. $H^l$ and $W^l$ indicate the height and width of channels in the $l$th layer, respectively. Pruning $k$th channel in layer $l$ means removing the whole filter $\mathbf{w}^l_k$. We further define $\mathbf{u}^l$ to represent all channel utilities in $l$th layer and $\mathbf{u}^l\in \mathbb{R}^{C^l}$. The forward channels are determined by channels utilities, which will be illustrated in the following sections.
\subsection{Forward channel selection}
As mentioned before, our proposed training scheme dynamically select channels to participate in the training through comparing their utility values. We retain the activation of channels with large utility values while zeroing out others. Given a pruning rate $p$ and the whole channel number $N$, we need to find a global threshold that determines the choice of channels with $pN$ channels masked to be zero. Let the notation $\mathbf{t}$ represent such threshold and we have
\begin{equation}
\label{e0.1}
\mathbf{t}\gets\text{p-min}_u\left\{u\in \mathbf{u}^l, 1\le l\le L\right\}
\end{equation}
where the function "$\text{p-min}$" returns the utility with the $pN$-th smallest value in all utility values. Next, we construct a binary mask $\mathbf{m}^l$ with initialized values of 0. By setting the specific values which are lager than the threshold $\mathbf{t}$ to 1, we have
\begin{equation}
\label{e0.2}
m^l_k={}
\begin{cases}
~~1,~~ u_k^l > \mathbf{t}\\
~~0,~~ u_k^l\le \mathbf{t}
\end{cases}
\end{equation}
we then multiply the channels by the constructed mask
\begin{equation}
\label{e0.3}
\mathbf{z}^l_{k} \gets \mathbf{z}^l_{k} \cdot m^l_k,~1\le k\le C^l 
\end{equation}

After doing so, the activation values of channels with respect to lower utility values are all set to 0. In fact, the key of our proposed scheme is to pick up the significant channels through the construction of the mask $\mathbf{m}^l$, which is stemed from (\ref{e0.1}). The threshold $\mathbf{t}$ needs to be updated as long as the channel utility $\mathbf{u}^l$ updates. 
\subsection{Criterion with channel significance}
In this section, we introduce the criterion on evaluating the significance of channel. We adopt the same criterion derived from Taylor expansion \cite{molchanov17}. Considering a mini-batch $B=\left\{X=\{x_1, x_2,..., x_m\}, Y=\{y_1, y_2, ..., y_m\}\right\}$, the final loss on the batch $B$ can be defined as $J(B,W)$ where $W$ represents the network parameters. If the kernel $\mathbf{w}^l_k$ with respect to the activation $\mathbf{z}^l_k$ is removed, the change in the cost function can be expressed as
\begin{equation}
\label{e1}
\left|\Delta J(\mathbf{z}^l_k)\right| = \left|J(B,\mathbf{z}^l_k) - J(B, \mathbf{z}^l_k\to 0)\right|
\end{equation}

We use the Taylor series to expand the cost function at point $\mathbf{z}^l_k=0$, as
\begin{equation}
\label{e2}
J(B, \mathbf{z}^l_k\to 0) = J(B, \mathbf{z}^l_k) - \frac{\partial J}{\partial \mathbf{z}^l_k}\mathbf{z}^l_k + o\left(\left(\mathbf{z}^l_k\right)^2\right)
\end{equation}

Ignoring the higher-order remainder and substituting (\ref{e2}) to (\ref{e1}), we have
\begin{equation}
\label{e3}
\begin{split}
\Theta_k^l \triangleq \left|\Delta J(\mathbf{z}^l_k)\right|
&= \left|J(B,\mathbf{z}^l_k) - J(B, \mathbf{z}^l_k) + \frac{\partial J}{\partial \mathbf{z}^l_k}\mathbf{z}^l_k\right| \\
&= \left|\frac{\partial J}{\partial \mathbf{z}^l_k}\mathbf{z}^l_k\right|
\end{split}
\end{equation}

The criterion can be seen as a measure of the significance of feature maps. For a channel with multi-variate output, the item $\Theta^l_k$ can be rewritten as
\begin{equation}
\label{e4}
\Theta_k^l = \left|\frac{1}{M}\sum_{m=1}^{M}\frac{\partial J}{\partial z^l_{k, m}} z^l_{k, m}\right|
\end{equation}
where $M$ is the total number of channel's entries. The computation of item $\Theta_k^l$ requires the activation and the gradient, which can be calculated from the forward and backward propagation, respectively. We impose an extra re-scaling method with max-normalization, that is
\begin{equation}
\label{E2}
\hat{\Theta}_k^l = \frac{\Theta_k^l}{\max\limits_{j}\left\{\Theta_j^l\right\}}
\end{equation}

Such normalization is essential as we need to ensure that the channel utility of each layer is at the same scale. The equation (\ref{E2}) shows that the maximum criterion values of each layer are all normalized to 1, resulting in similar channel-utility scales for different layers. 

It is worth noting that this criterion is evaluated on a single mini-batch. While the original work \cite{molchanov17} suggests to accumulate the criterion over the entire training period to evaluate the saliency of channels, we propose to utilize the criterion after each iteration and directly pick up the significant channels during training.

The final step in our approach is to accumulate the normalized criterion into the channel utility, as
\begin{equation}
\label{E6}
\left[u^l_k\right]_{\text{new}}\gets \lambda\left[u^l_k\right]_{\text{old}} + \hat{\Theta}_k^l
\end{equation}
where $\lambda$ is the decay factor. The motivation of process (\ref{E6}) is to realize dynamical channel selection as introduced before. In other words, if the updated amount of the channel utility is less than the amount of attenuation, the channel utility will decrease and the corresponding channel may not be selected in the next iteration. Of note is that the decay factor will drop as the learning rate decreases, which implies that the process of channel selection in forward propagation will stop in the final period of training to ensure the convergence.

We summarize the above process in Algorithm 1 where the forward channels are chosen by constructing a binary mask $\mathbf{m}^l$ whose elements indicate whether the corresponding channels are kept or not. The product of binary mask $\mathbf{m}^l$ and channels $\mathbf{z}^l$ is taken in a entry-wise manner that every entry in $\mathbf{m}^l$ with zero value zeros all the values in the corresponding channel in $\mathbf{z}^l$. For the reason that channel pruning is only applied to convolutional layers, we have omitted the general batch normalization \cite{serpy2015batch} layers, activation layers, pooling layers, and fully connected layer for simplicity.
\begin{algorithm}[t]
	\caption{Dynamic Channel Propagation} 
	{\bf Input:} 
	input image $\mathbf{z}^0$, pruning rate $p$, model $\left\{\text{conv}_l, 1\le l \le L\right\}$, decay factor $\lambda$, maximize iteration $I_{max}$\\
	{\bf Output:} 
	a more compact model
	\begin{algorithmic}
		\FOR{iteration $i\le I_{max}$, $i \leftarrow i+1$}
		\STATE forward propagation 
		\FOR{$l=1$ to $L$} 
		\STATE calculate the output channels ~~~$\mathbf{z}^l \leftarrow \text{conv}_l(\mathbf{z}^{l-1})$
		\STATE generate a mask $\mathbf{m}^l$ initialized by zero~~~$\mathbf{m}^l \leftarrow \mathbf{0}$
		\STATE find the global threshold $\mathbf{t}$ and set the corresponding $m^l_i$ to 1 refer to (\ref{e0.1}).~~~$m^l_i\gets 1$ if $u_k^l > \mathbf{t}$
		\STATE multiply channels by the mask~~~$\mathbf{z}^l \leftarrow \mathbf{z}^l \cdot \mathbf{m}^l$
		\ENDFOR
		\STATE calculate the final loss $J$ 
		\STATE set $\mathbf{g}^{L+1}\leftarrow \mathbf{1}$ and $\mathbf{z}^{L+1} \leftarrow J$
		\STATE backward propagation
		\FOR{$l=L$ to $1$}
		\STATE calculate the channel gradients~~~$\mathbf{g}^l \leftarrow \mathbf{g}^{l+1}\cdot\frac{\partial \mathbf{z}^{l+1}}{\partial \mathbf{z}^l}$
		\STATE compute the criterion described in (\ref{e4}) and (\ref{E2})
		\STATE ~~~~~$ \hat{\Theta}_k^l \leftarrow \text{Max-norm}\left[\text{average}\left(\mathbf{g}^l \cdot \mathbf{z}^l\right)\right]$
		\STATE update the channel utility~~~$u^l_k\gets \lambda u^l_k + \hat{\Theta}_k^l$
		\ENDFOR
		\ENDFOR
		\STATE prune channels and the corresponding filters according to $\mathbf{u}^l$ in each layer $l$
		\STATE copy the remaining parameters to a compact model
		\RETURN the compact model
	\end{algorithmic}
\end{algorithm}
\subsection{Discussions}
Notice that our approach does not need the time-consuming re-training process which is the required fundamental step of the previous pruning approaches. The proposed scheme only adjusts those "valid" channels with high channel utility and the corresponding convolutional kernels in the training time. Therefore, after the training is complete, we only need to copy the convolutional kernel parameters with respect to those valid channels to a more compact model.

Our dynamic training scheme can also begin with a well-trained model. In other words, if given a well-trained but redundant model, our method can turn it into a more compact one directly. Such process is similar to fine-tuning. Previous pruning approaches \cite{liu2017learning}, \cite{wen2016Learning} introduce additional loss to the training phase and they must formally train a large model to learn the saliency of channels (or the sparsity) on the basis of their introduced loss. However, if given a well-trained model, they do not have the criteria on which pruning depends and have to train it again to obtain the sparsity of such model.

As most of the computation is concentrated on the convolutional layers, we only need to pay attention to the computational savings of these layers. Suppose the output channel size of the $l$th layer is $H^l\times W^l$ and the pruning rate is $p^l$, which indicates that the corresponding $p^l\times C^l$ filters will be removed after training. Therefore, the dimension of remaining channels in the $l$th layer is $H^l\times W^l \times (1-p^l)\times C^l$ and the overall computation (FLOPs) decreases from $K^2\times C^{l-1}\times H^l\times W^l \times C^l$ to $K^2\times C^{l-1}\times H^l\times W^l \times (1-p^l)\times C^l$, where the label $K$ indicates kernel size. While considering the pruned channels in the previous layer, the final FLOPs in the $l$th layer is $K^2\times (1-p^{l-1})\times C^{l-1}\times H^l\times W^l \times (1-p^l)\times C^l$. Compared to the raw FLOPs, a proportion of $1-(1-p^{l-1})(1-p^l)$ is reduced, which leads to significant computational savings and reduced network inference time.
\section{Experiments}
\subsection{Settings}
We evaluate our scheme on various representative benchmark datasets, including CIFAR-10 \cite{alex2009learning}
and ILSVRC-12 \cite{jia2009imagenet}, and typical advanced network architectures, including VGG \cite{szegedy2015going} and ResNet \cite{he2016residual}. CIFAR-10 contains 50k training images and 10k testing images, which are categorized into 10 classes. We follow the common data augmentation suggested by  \cite{he2016residual} with shifting and mirroring. Both VGG and ResNet are trained using Stochastic Gradient Descent (SGD) with an initial learning rate of 0.1 which is decayed by a factor of 0.1 in one-third and two-third of the total epochs. The weight decay and momentum is $10^{-4}$ and 0.9, respectively. As mentioned before, the decay constant $\lambda$ is initialized by 0.6 and divided by 10 along with the learning rate decaying to ensure the convergence. When evaluating on CIFAR-10, we both train the model from scratch and from a post-training initialization to show the effectiveness of our approach. ILSVRC-12 contains 1.3 million training images and 50k validating images without test set. While experimenting on ILSVRC-12, we follow the training settings and the strategy of data augmentation suggested by \cite{he2016residual}. Our implement is based on the framework PyTorch \cite{adam2017pytorch}.

We compare our proposed scheme with various state-of-the-art pruning approaches, including PEFC \cite{li2017pruning}, NS \cite{liu2017learning}, CP \cite{he2017channel}, ThiNet \cite{luo2017ThiNet}, SFP \cite{he2018soft}, Leverage \cite{singh2018leveraging}, DCP \cite{zhuang2019discrimination}, FPGM \cite{he2019fpgm}, and COP \cite{Wang2019cop}. Overall, our proposed scheme has achieved comparable results with simpler pruning procedure. 
\subsection{VGG on CIFAR-10}
We adopt a 16-layer VGG structure, including 13 convolutional layers. The channels in each layer are [64, 64, 128, 128, 256, 256, 256, 512, 512, 512, 512, 512, 512]. We have run several experiments where the pruning FLOPs differ. All the results are summarized as Table \ref{table1}. 
\begin{table}[t]
	\centering
	\caption{Results of pruning VGG on CIFAR-10. We report both baseline and after-pruned accuracy of the state-of-the-art methods refer to their papers. The headline "Re-trained" indicates that whether the results are obtained on the basis of the traditional three-stage process or not.}
	\begin{tabular}{ccccc}
		\toprule
		\multirow{2}{*}{Method}&Re-&\multirow{2}{*}{Base.(\%)}&\multirow{2}{*}{Acc.(\%)}&FLOPs\\
		&trained?&& &pruned(\%)\\
		\midrule
		\multirow{3}{*}{Ours} &  \multirow{3}{*}{\xmark}&  \multirow{3}{*}{93.50}&$\uparrow$ 0.29& 34.1\\
		& &	&$\uparrow$ 0.10& 51.9\\
		&  & &$\downarrow$ 0.50& 73.3\\
		\midrule
		PFEC&\cmark&93.25&$\uparrow$ 0.15&34.2\\
		\midrule
		NS&\cmark&93.66&$\uparrow$ 0.14&51.0\\
		\midrule
		DCP&\cmark&93.99&$\uparrow$ 0.17& 50.0\\
		\midrule
		\multirow{3}{*}{Ours} &  \multirow{3}{*}{\cmark}&  \multirow{3}{*}{93.50}&$\uparrow$ 0.30& 34.1\\
		& &	&$\uparrow$ 0.15& 51.9\\
		&  & &$\downarrow$ 0.51& 73.3\\
		\bottomrule
	\end{tabular}
	\label{table1}
\end{table}

As shown in Table \ref{table1}, we run several experiments which are initialized with and without well-trained models. In case of training the model from well-trained models, our proposed scheme can achieve comparable performance with the aforementioned state-of-the-art methods in various pruned FLOPs, realizing the compact model with 51.9\% in FLOPs drop but whose accuracy is even higher compared to the baseline. We also find that our scheme can achieve competitive results if train the model from scratch, as shown in the first three rows in Table \ref{table1}.
\subsection{ResNet on CIFAR-10}
We adopt the 32-layer ResNet structure recommended in the article \cite{he2016residual}. ResNet-32 contains three stages and each stage includes 10 convolutional layers. The numbers of output channels in each stage are 16, 32 and 64. Compared to VGG, ResNet-32 is more compact because it contains fewer channels in each layer and therefore is more challenging to prune. We run several experiments where the pruning rate varies and summarize the results as Table \ref{table2}. 
\begin{table}[b]
	\centering
	\caption{Results of pruning ResNet-32 on CIFAR-10. The headings are with the same meaning as Table \ref{table1}.}
	\begin{tabular}{ccccc}
		\toprule
		\multirow{2}{*}{Method}&Re-&\multirow{2}{*}{Base.(\%)}& \multirow{2}{*}{Acc.(\%)}&FLOPs\\
		&trained?&&  &pruned(\%)\\
		\midrule
		SFP&\xmark&92.63&$\downarrow$ 0.55&41.5\\
		\midrule
		FPGM&\xmark&92.63&$\downarrow$ 0.70&53.2\\
		\midrule
		\multirow{2}{*}{Ours} &  \multirow{2}{*}{\xmark}&  \multirow{2}{*}{93.30}&$\downarrow$ 0.44& 41.4\\
		&&&$\downarrow$ 0.65 & 50.2\\
		\midrule
		COP&\cmark&92.64&$\downarrow$ 0.67&53.9\\
		\midrule
		\multirow{2}{*}{Ours} &  \multirow{2}{*}{\cmark}&  \multirow{2}{*}{93.30}&$\downarrow$ 0.40& 41.4\\
		&  & &$\downarrow$ 0.70& 50.2\\
		\bottomrule
	\end{tabular}
	\label{table2}
\end{table}

While pruning ResNet-32, our proposed scheme can achieve competitive results, such as an accuracy of 92.90\% with nearly 41.4\% reduction in FLOPs, whose accuracy outperforms that of SFP. While pruning more channels of the original model, our scheme can still reach a competitive performance with only 0.70\% drop in accuracy but 50.2\% in FLOPs reduction.
\subsection{ResNet on ILSVRC-12}
We adopt a widely studied architecture ResNet-50 as in many pruning approaches. Different from general ResNet architecture, ResNet-50 contains a special structure called "bottleneck" \cite{he2016residual}, including three convolutional layers that only the middle layer is expressive in each residual block. In our experiments, we focus on pruning the channels of the first two layers in a bottleneck whose benefit is that we do not need to worry about the identity mapping when copying the parameters to a compact model since the input/output dimensions within the same residual block do not change. 

We summarize the experimental results on ILSVRC-12 as in Table \ref{table4}, where one can see that our scheme can reduce the FLOP calculation of the model by 41.1\%, but still maintains a competitive accuracy of 74.25\% and 92.05\% in top-1 and top-5, respectively. This result has exceeded that of \cite{luo2017ThiNet} and there is no fine-tuning. In the case of the pruned FLOPs increasing to 52.7\%, our approach can still achieve 73.32\% and 91.65\% in top-1 and top-5 accuracy, respectively.   
\begin{table*}[t]
	\centering
	\caption{Results of pruning ResNet-50 on ILSVRC-12.}
	\begin{tabular}{ccccccc}
		\toprule
		\multirow{1}{*}{Method} & \multirow{1}{*}{Re-trained?}&Top-1 baseline(\%)&\multirow{1}{*}{Accuracy} &Top-5 baseline(\%)&\multirow{1}{*}{Accuracy}&FLOPs pruned(\%)\\
		\midrule
		\multirow{2}{*}{ThiNet}&\multirow{2}{*}{\cmark}& \multirow{2}{*}{72.88}&$\downarrow$ 0.84& \multirow{2}{*}{91.14}&$\downarrow$ 0.47&36.8\\
		&&&$\downarrow$ 1.87&&$\downarrow$ 1.12&55.8\\
		\midrule
		CP&\cmark&-&-&92.20&$\downarrow$ 1.40&$\approx$ 50.0\\
		\midrule
		\multirow{2}{*}{SFP}&\xmark& \multirow{2}{*}{76.15}&$\downarrow$ 1.54& \multirow{2}{*}{92.87}&$\downarrow$ 0.81&\multirow{2}{*}{41.8}\\
		&\cmark&&$\downarrow$ 14.01&&$\downarrow$ 8.27&\\
		\midrule
		Leverage&\cmark&75.30&$\downarrow$ 1.90&92.20&$\downarrow$ 0.80&$\approx$ 50.0\\
		\midrule
		\multirow{2}{*}{Ours}&\multirow{2}{*}{\xmark}& \multirow{2}{*}{76.13}&$\downarrow$ 1.88& \multirow{2}{*}{92.86}&$\downarrow$ 0.81&41.1\\
		&&&$\downarrow$ 2.81&&$\downarrow$ 1.21&52.7\\
		\bottomrule
	\end{tabular}
	\label{table4}
\end{table*}
\section{Ablation Study}
\subsection{Various pruning rates}
\begin{figure}[t]
	\centering
	\includegraphics[scale=0.52]{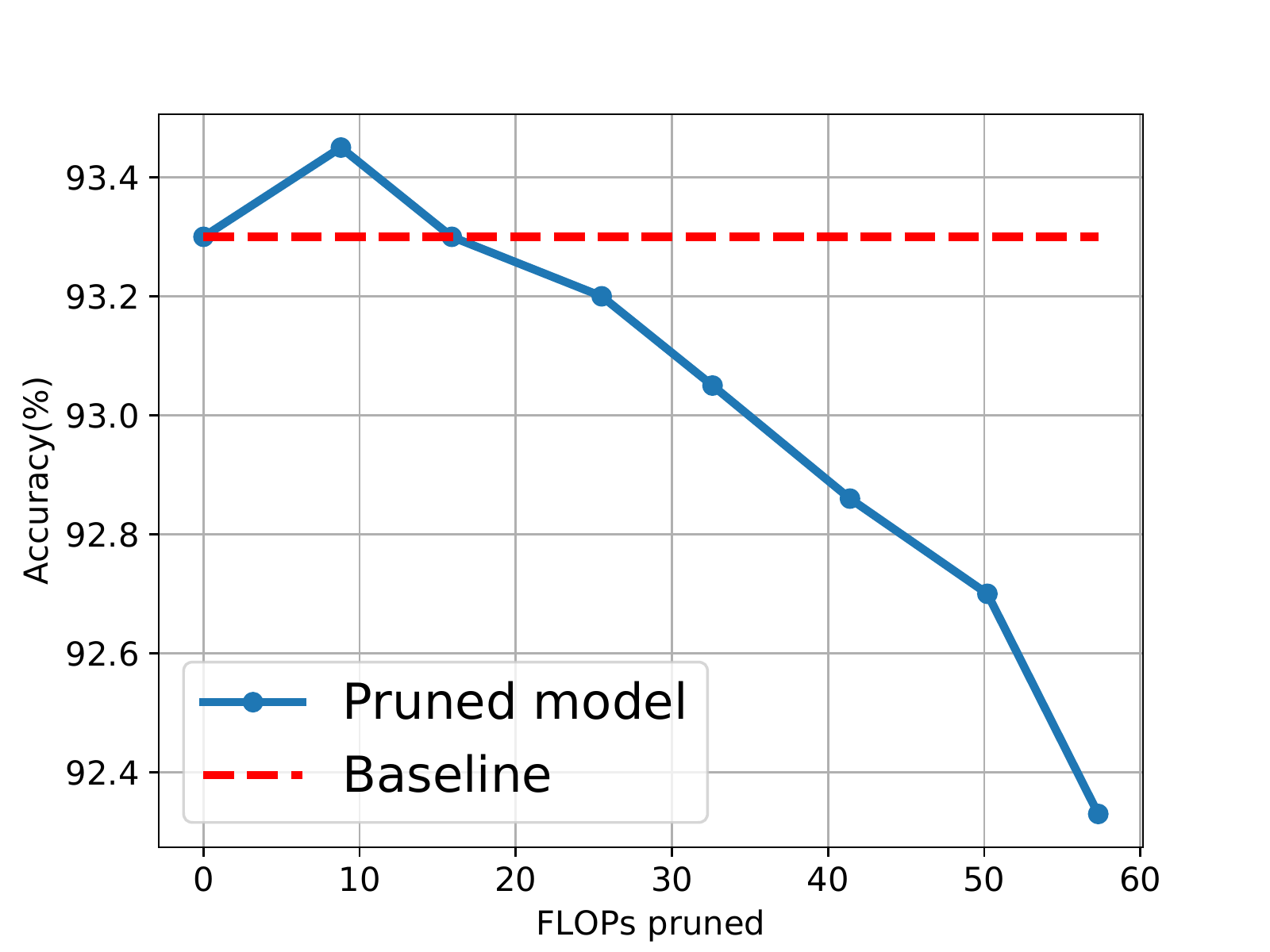} 
	\caption{The pruned results with respect to various pruning rates. The experiments are performed on CIFAR-10 with ResNet-32.}
	\label{f3}
\end{figure}

\begin{figure}[t]
	\centering
	\includegraphics[scale=0.52]{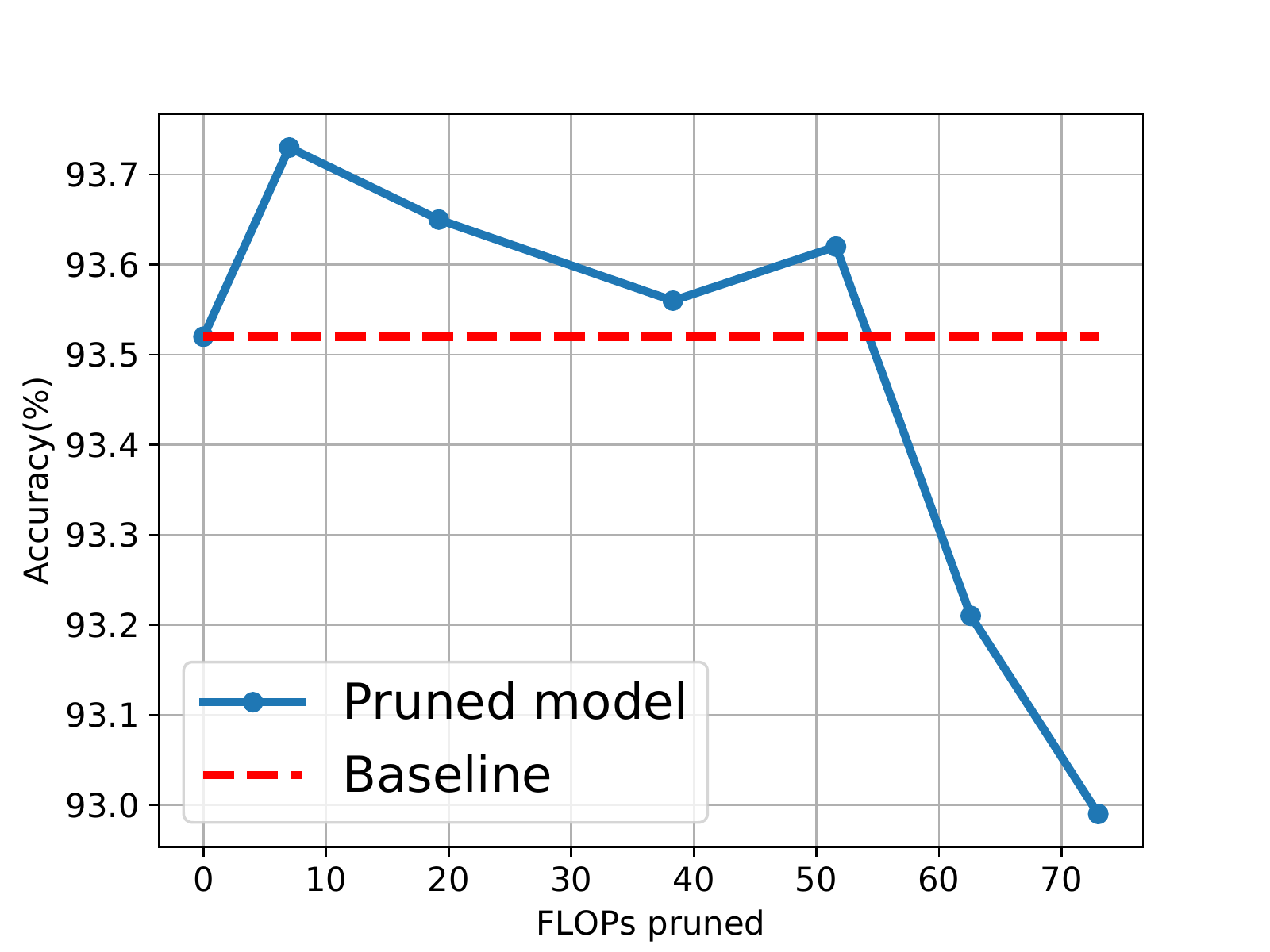} 
	\caption{The pruned results with respect to various pruning rates, which is obtained from pruning VGG-16 on CIFAR-10.}
	\label{f4}
\end{figure}

In this section, we explore the performance of our scheme with regard to different pruning rates. For all experiments, we use the same hyper-parameter settings and train the model from scratch. We plot the results as Fig.\ref{f3}. As depicted in Fig.\ref{f3}, our scheme has almost no performance loss at low compression rates when pruning ResNet. In the case when the pruned FLOPs increase to 50\%, our scheme still has an ideal accuracy of 92.65\%, which is qualified in such data set. We also summarize the accuracy with respect to various pruned FLOPs on VGG. As depicted in Fig.\ref{f4}, our scheme can learn efficient architectures with even higher testing accuracies compared to their baseline at the low level of pruned FLOPs. While the pruned FLOPs is increased to 70\%, our proposed scheme can maintain a competitive accuracy up to 93.00\%.  
\subsection{Decay factor}
As mentioned before, our proposed algorithm has a specific hyper-parameter, namely the decay factor $\lambda$. If this parameter is set to a small value, the channel utility may fluctuate greatly and the training may be unstable. Conversely, if this value is too large, there is no effect of dynamic channel propagation. We adopt the strategy that the decay factor increases at the same time when the learning rate decreases in the training. We compare the results with that of fixed decay factor and summarize the corresponding results in Table \ref{table5}.
\begin{table}[t]
	\centering
	\caption{The comparison of accuracy regarding different values of $\lambda$ or updating strategies.}
	\begin{tabular}{ccccc}
		\toprule
		\multirow{2}{*}{Architecture}&\multicolumn{3}{c}{fixed $\lambda$ (\%)}&mutative\\
		&0.1&0.5&0.9&strategy(\%)\\
		\midrule
		VGG&93.45&93.40&93.15&\textbf{93.71}\\
		\midrule
		ResNet&92.43&92.56&92.65&\textbf{92.90}\\
		\bottomrule
	\end{tabular}
	\label{table5}
\end{table}

As can be observed from Table \ref{table5}, the changeable strategy of decay factor can retain better performance compared to the fixed strategy.
\subsection{The architectures after pruning}
\begin{figure}[t]
	\centering
	\includegraphics[scale=0.55]{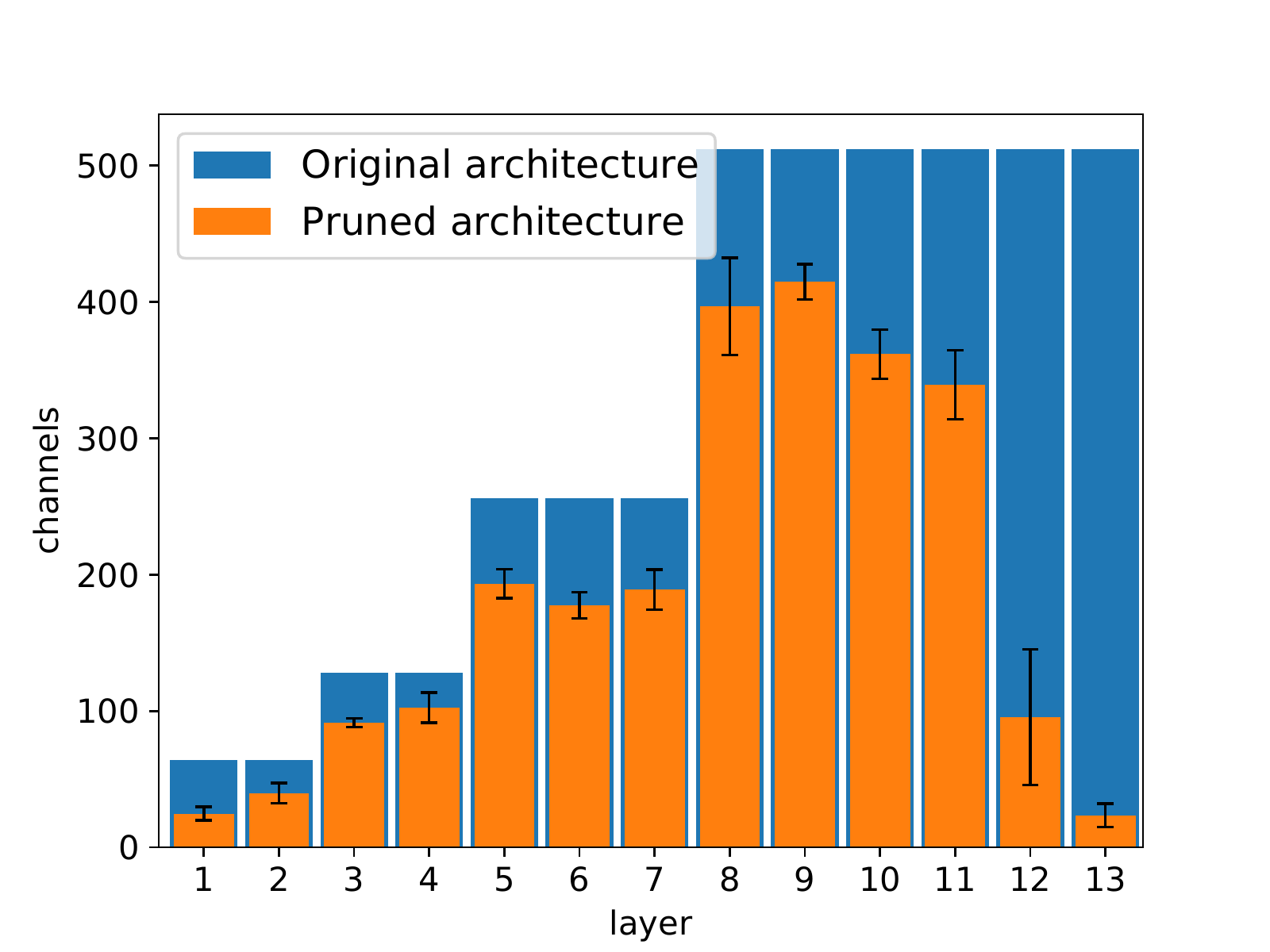} 
	\caption{The channel distribution of the pruned model. The result is obtained by pruning 40\% channels of VGG network.}
	\label{f5}
\end{figure}

\begin{figure}[t]
	\centering
	\includegraphics[scale=0.55]{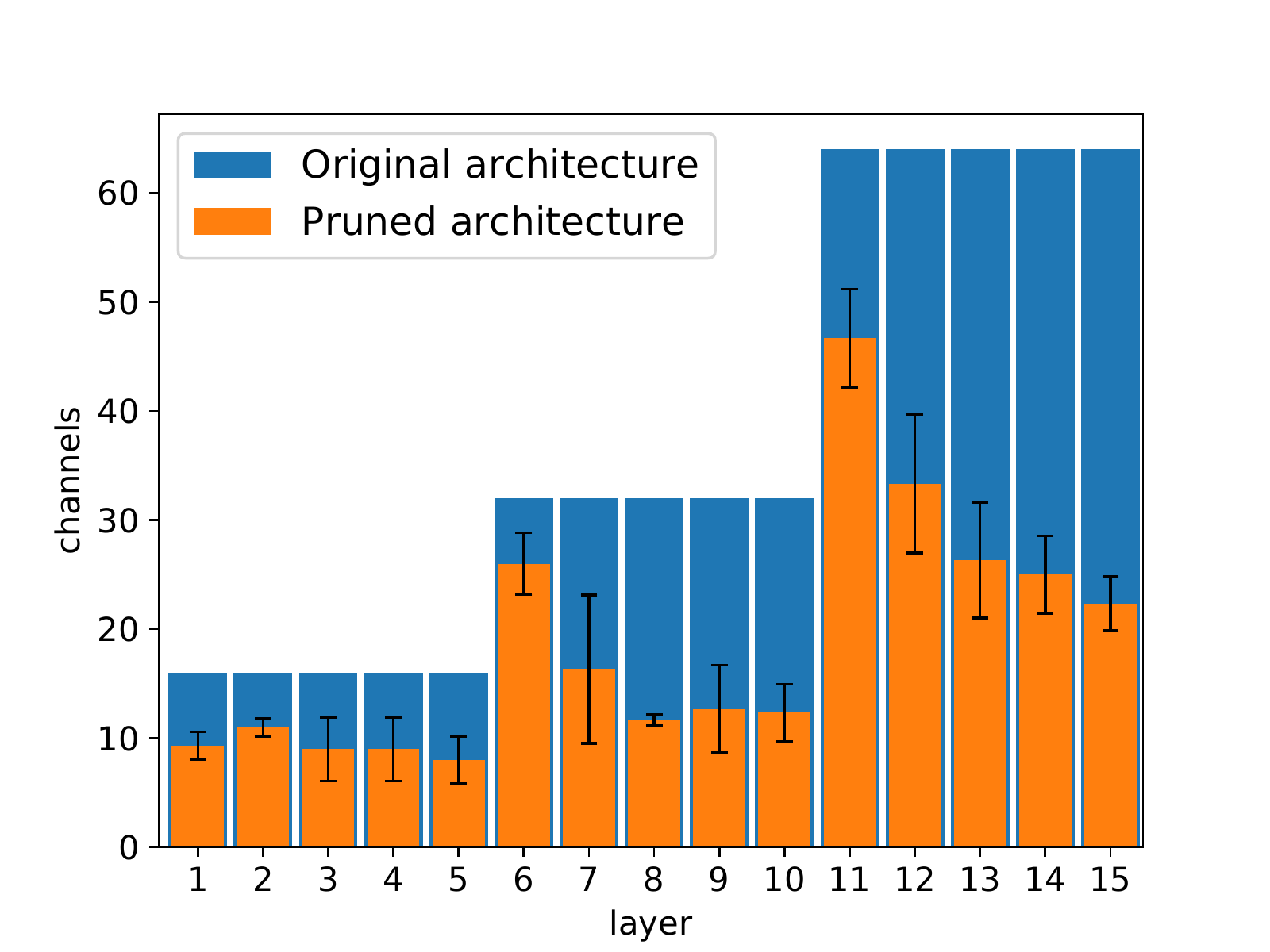} 
	\caption{The channel distribution of the pruned model for ResNet architecture. The result is obtained by pruning 50\% channels of the first layer within each residual block.}
	\label{f6}
\end{figure}

In the section, we show the sub-network architectures obtained by our proposed scheme. As can be observed from Fig.\ref{f5}, our scheme keeps more channels in the middle layers, while pruning more channels in the later layers and the first layer. The unveiled structure suggests that middle layers are more sensitive whereas the first layer and later layers are easier to be pruned. Such conclusion is consistent with the previous findings of \cite{li2017pruning, liu2017learning}, which indicates the effectiveness of our algorithm. Fig.\ref{f6} shows the result on ResNet with 50\% channels removed from the original network. The pruned structure indicates that the layers in which the channel number doubles seem to retain more channels (the 5th layer and the 10th layer in Fig.6), which is consistent with the finding in \cite{li2017pruning} that the channels where the channel size changes are more sensitive.

\section{Conclusion}
In this paper, we present a novel neural network training algorithm that picks up the most important channels from the redundant convolutional layers and simplify the general channel-pruning procedure by selectively adjusting the parameters of the critical kernel filters in the training phase. Our scheme is light-weight and can be easily incorporated into traditional training process of neural networks. In addition, the proposed scheme can determine a good sub-architecture as long as given a compression rate. A large amount of experiments based on a number of benchmark datasets verify the effectiveness of our algorithm.
\section*{Acknowledgement}
This work was supported in part by National Key R\&D Program of China (No. 2017YFB1301003), National Natural Science Foundation of China (No. 61701439, 61731002), Zhejiang Key Research and Development Plan (No. 2019C01002, 2019C03131), the Project sponsored by Zhejiang Lab (2019LC0AB01), Zhejiang Provincial Natural Science Foundation of China (No. LY20F010016).

\bibliographystyle{IEEEtran}
\bibliography{ref}

\end{document}